\documentclass[mlabstract]{jmlr}

\usepackage{listings}
\usepackage{xcolor}

\definecolor{codegreen}{rgb}{0,0.6,0}
\definecolor{codegray}{rgb}{0.5,0.5,0.5}
\definecolor{codepurple}{rgb}{0.58,0,0.82}
\definecolor{backcolour}{rgb}{0.95,0.95,0.92}

\lstdefinestyle{mystyle}{
    backgroundcolor=\color{backcolour},   
    commentstyle=\color{codegreen},
    keywordstyle=\color{magenta},
    numberstyle=\tiny\color{codegray},
    stringstyle=\color{codepurple},
    basicstyle=\ttfamily\footnotesize,
    breakatwhitespace=false,         
    breaklines=true,                 
    captionpos=b,                    
    keepspaces=true,                 
    numbers=left,                    
    numbersep=5pt,                  
    showspaces=false,                
    showstringspaces=false,
    showtabs=false,                  
    tabsize=2
}

\usepackage{longtable}

\usepackage{booktabs}
\usepackage[load-configurations=version-1]{siunitx} 
\usepackage{fontawesome5} 
\usepackage{hyperref}

\theorembodyfont{\upshape}
\theoremheaderfont{\scshape}
\theorempostheader{:}
\theoremsep{\newline}

\jmlrvolume{}
\firstpageno{1}

\jmlryear{2026}
\jmlrworkshop{Symmetry and Geometry in Neural Representations}

\title[REViT-v2]{REViT-v2: Hierarchical Windowed Roto-reflection Equivariant ViT
for Equivariant Feature Extraction\titletag{}} 

 \author{\Name{Sheir A. Zaheer} \Email{sheir@kc-ml2.com}\\
 \addr KC Machine Learning Lab,  Rep. of Korea
 \AND
 \Name{Jihwan Moon\nametag{\thanks{work done at KC Machine Learning Lab}}} \Email{jimoon1215@gmail.com}\\
 \addr {NFOCZ Inc \and Seoul National University, Rep. of Korea}
 \AND
 \Name{Chan Y. Park} \Email{chan.y.park@kc-ml2.com}\\
 \addr KC Machine Learning Lab,  Rep. of Korea
 }

\begin{document}

\maketitle

\begin{abstract}
We propose a scalable roto-reflection-group-equivariant vision transformer based on windowed group-convolutional self-attention and a hierarchical feature architecture. We demonstrate that our approach can be scaled to group-equivariant vision transformers (ViTs) with millions of parameters and large datasets with practically sized images, i.e., ImageNet. The code and pretrained weights for the proposed Hierarchical Windowed Roto-reflection Equivariant ViTs (REViT-v2)  are available at \faGithub{} \url{https://github.com/kc-ml2/revit}.
\end{abstract}

\section{Introduction}
\label{sec:intro}

Standard vision transformers (ViTs) \citep{dosovitskiy2021image} do not explicitly encode common image symmetries. A conventional model must learn from data that a visual pattern retains its identity after a rotation or reflection. Group-equivariant neural networks instead incorporate this relationship into the architecture. 
Group-equivariant convolutional networks introduced group convolutions as a systematic means of sharing filters across translations, rotations, and reflections \citep{cohen2016group, weiler2019general}. 

\cite{romero2021group} extended the same principle to attention and showed that attention can be made equivariant to symmetry groups by constructing positional encodings that are invariant under the corresponding group action. This provides a general theoretical formulation, but it does not scale beyond low-resolution images. Moreover, group-relative positional terms must be repeatedly evaluated in all attention layers. \cite{zaheer2026revit} subsequently introduced rot-reflection equivariant convolutional vision transformer (REViT) with group-convolutional self-attention (G-CSA), in which convolutional projections remove the dependence on explicit positional encoding. This makes equivariant transformers considerably simpler. Nevertheless, when group-convolutional attention is evaluated globally, its spatial attention matrix remains quadratic with respect to input size. The cost is especially important for equivariant models because regular group representations typically allocate several orientation channels to each feature field. 

Several developments in efficient vision transformers suggest a path toward addressing this limitation. CvT introduced convolutional token embedding and convolutional attention projections showing that convolutional structure can reduce the need for explicit positional encoding \citep{wu2021cvt}. Swin transformer restricted attention to local windows and organized the network into a multiscale pyramid \citep{liu2021swin}. Multiscale Vision Transformers similarly demonstrated the value of progressively reducing token resolution while increasing channel capacity \citep{fan2021multiscale}. 

In this paper, we propose REViT-v2. REViT-v2 combines hierarchical vision transformers with local group-equivariant convolutional self attention. REViT-v2 introduces two central components. First, it utilizes Windowed Group Convolutional Self-Attention (wG-CSA), which performs attention within local spatial windows while preserving complete group-representation fields in each attention head. Second, it constructs a multi-stage equivariant feature pyramid using group-convolutional tokenization and downsampling.  Consequently, REViT-v2 reduces the spatial complexity of group-equivariant attention from quadratic to linear in the image area. This design targets large-scale image recognition while maintaining equivariance throughout the feature extractor. We demonstrate its effectiveness through classification results on ImageNet-1K \citep{imagenet} and a complexity comparison with global G-CSA.

\section{Windowed Group Convolutional Self-Attention}
\label{sec:windowed_attention}
wG-CSA is a local window extension of G-CSA \citep{zaheer2026revit}. The aim is to reduce the quadratic complexity of G-CSA in image size to linear complexity.    


\subsection{Equivariant Projections}
Let the input to an attention block be a feature map. The channels of input $x$ are organized according to a representation $\rho$ of a finite planar symmetry group $G$, such as the cyclic rotation group $pN$ or the dihedral group $pNm$. 
Queries, keys, and values are produced by group-equivariant convolutional projections \cite{zaheer2025group}:
\begin{equation}
    Q=\phi_Q(x),\qquad K=\phi_K(x),\qquad V=\phi_V(x),
    \label{eq:qkv}
\end{equation}
The attention output is followed by another group-equivariant projection $\phi_O$.


\subsection{Local Window Attention}

The spatial lattice is partitioned into non-overlapping windows,
    $\mathcal{W}=\{W_1,\ldots,W_K\}$,
where each window contains $M^2$ positions. Let $W(p)$ denote the window containing position $p$. For head $h$, the attention score between $p$ and $q\in W(p)$ is
\begin{equation}
    s_{pq}^{h}=
    \frac{\left\langle Q^{h}(p),K^{h}(q)\right\rangle}{\sqrt{d_h}},
    \label{eq:attention_score}
\end{equation}
where $Q^h$ and $K^h$ are query and key projections for head $h$.
The normalized attention coefficient and output are
\begin{align}
    a_{pq}^{h}
    &=
    \frac{\exp(s_{pq}^{h})}
    {\sum_{r\in W(p)}\exp(s_{pr}^{h})},
    \\
    Y^{h}(p)
    &=
    \sum_{q\in W(p)}a_{pq}^{h}V^{h}(q),
    \label{eq:attention_output}
\end{align}
where $V^h$ is the value projection.
Outputs from all heads are concatenated along complete representation fields and projected using $\phi_O$:
\begin{equation}
    \operatorname{wG-CSA}(x)=
    \phi_O\!\left(\operatorname*{Concat}_{h=1}^{H}Y^h\right).
    \label{eq:wgcsa}
\end{equation}
No absolute positional embedding is introduced. Spatial structure is instead preserved by the equivariant convolutional stem and the convolutional query, key, and value projections.
We demonstrate the equivariance of  wG-CSA in \emph{Appendix} \ref{apd:eq}.

\subsection{Linear Complexity of Windowed G-CSA}

For $x$ of size $H \times W$, $N=HW$ denote the number of positions in a feature map. Global attention computes $N^2$ pairwise scores per head and has approximate complexity
    $O(N^2C)$ \citep{dosovitskiy2021image}.
Windowed attention with $M \times M$ sized windows, however, contains $N/M^2$ windows, each requiring $M^4$ pairwise comparisons. Its total complexity is therefore
\begin{equation}
    O\!\left( (N/M^2) M^4C\right)=O(NM^2C).
    \label{eq:window_complexity}
\end{equation}
For fixed $M$, the cost is linear in image area. 

\section{REViT-v2}
\label{sec:hierarchical_model}

REViT-v2 arranges the proposed wG-CSA in a multi-stage feature hierarchy. The early layers need to preserve sufficient spatial resolution to identify local structures, whereas the deeper layers need greater representation capacity and wider effective receptive fields. 

\subsection{Equivariant Patching and Lifting}

The input image is interpreted as a field of trivial representations because rotating an image changes spatial positions but does not permute its RGB channels. A group-equivariant convolutional stem lifts this input into regular group representations. Two strided group-equivariant convolutions reduce the image resolution by a factor of four.

\subsection{Equivariant Transformer Block}

Each block contains a pre-normalized wG-CSA module and an equivariant feed-forward network. Given input $x_l$, the block is
\begin{align}
    z_l
    &=x_l+
    \operatorname{DropPath}\!\left(
    \alpha_l\operatorname{wG-CSA}(\mathcal{N}_1(x_l))
    \right),
    \\
    x_{l+1}
    &=z_l+
    \operatorname{DropPath}\!\left(
    \beta_l\mathcal{M}_G(\mathcal{N}_2(z_l))
    \right).
    \label{eq:transformer_block}
\end{align}

Here, $\mathcal{N}_1$ and $\mathcal{N}_2$ are representation-compatible normalization operations, while $\alpha_l$ and $\beta_l$ are trainable residual scaling parameters. The equivariant MLP $\mathcal{M}_G$ replaces ordinary fully connected layers with $1\times1$ group convolutions. $\operatorname{DropPath}$ 
implements stochastic depth. Stochastic depth regularizes the residual branches without independently dropping orientation channels, which would violate their representation structure.

\subsection{Multistage Hierarchy}

After the blocks in each of the first three stages, a strided group convolution reduces the spatial resolution while the number of representation fields is increased.
Later stages can therefore use more channels and attention heads without exceeding the cost.
At deeper stages, a fixed $M\times M$ window also corresponds to a progressively larger region of the original image. The final stage can therefore model high-level interactions over a substantial portion of the input without constructing a global high-resolution attention matrix.



%


The hierarchy and local windows address different parts of the scaling problem: windowing eliminates the quadratic dependence on the number of tokens, while downsampling prevents the high-resolution representation from being propagated through the complete network. Together, they provide a practical route toward symmetry-aware ViT backbones for ImageNet-scale classification and, with suitable interfaces, dense prediction tasks.

\section{Experimental Results}
Table \ref{tab:imageNet} compares the performance of a $p4m$-equivariant REViT-v2 with two baselines: a vanilla ViT trained with rotation and flip augmentations and a $p4m$-equivariant RE-ResNet model. REViT-v2 outperforms both. Additional details are provided in \emph{Appendix} \ref{app:revit_in}.

\begin{table}[tb]
\centering
\scriptsize
\caption{Performance Comparison on ImageNet-1K.}
\label{tab:imageNet}
\begin{tabular}{|l|c|c|c|}
\hline
\textbf{Model}          & \textbf{Top-1 Accuracy (\%)} & \textbf{Top-5 Accuracy (\%)} & \textbf{Params.} \\ \hline
ViT-S w/ aug   &   72.08    &   89.54    &  22 M       \\ \hline
RE-ResNet       &   77.37    &   93.74    & 11 M    \\ \hline
REViT-v2-T          &  72.58    &   90.88    &   5 M  \\ 
REViT-v2-S          &   {79.27}    &   {94.45}    & 18 M    \\
REViT-v2-S          &   {80.9}    &   {95.1}    & 47 M    \\ \hline

\end{tabular}%
\end{table}

\subsection*{Complexity Comparison with Global G-CSA}
Table \ref{table:compcomp} compares wG-CSA with global G-CSA on Rotated MNIST \citep{rotmnist} trained for 200 epochs and a batch size of 128. The proposed windowed architecture substantially reduces model size, forward FLOPs, latency, and peak memory while maintaining comparable accuracy. Since REViT-v2 combines a downsampling stem with hierarchical windowed attention, we also evaluate global attention after the same downsampling stem to distinguish the effect of early downsampling from that of windowed attention.
\begin{table}[tb]
\caption{Comparison with global G-CSA on Rotated MNIST.}
\label{table:compcomp}
\resizebox{\columnwidth}{!}{%
\begin{tabular}{|l|c|c|c|c|c|c|}
\hline
\textbf{Attention} & \textbf{Params.} & \textbf{Forward FLOPs} & \textbf{Latency vs global} & \textbf{Peak Training Memory}  & \textbf{Accuracy (\%)} \\ \hline
Global             & 97.7 K            & 1.69 GFLOPs            & 1$\times$                  & 429.7 MB              & 98.23             \\ \hline
Global w/downsampling     & 97.7 K            & 333.6 MFLOPs           & 0.75$\times$               & 50.98 MB              & 98.28                  \\ \hline
Windowed           & 102.9 K           & 80.1 MFLOPs            & 0.61$\times$               & 26.47 MB              & 98.26             \\ \hline
\end{tabular}%
}
\end{table}

\section{Conclusion}
In this paper, we demonstrate the effectiveness of REViT-v2 as scalable group-equivariant vision transformers that can serve as backbone feature extractors in the same manner as conventional ViTs. We also provide code and pretrained weights to support further research. 



\bibliography{pmlr-sample}

\appendix

\section{Equivariance Analysis}\label{apd:eq}

\subsection{Theoretical Analysis}

Assume that the group representations within a head are orthogonal, as is the case for permutation-based regular representations \citep{cesa2022a} used in this work:
\begin{equation}
    \rho_h(g)^{\top}\rho_h(g)=I.
\end{equation}
Under a transformation $g$, the projected features satisfy
\begin{equation}
    Q^{h'}(gp)=\rho_h(g)Q^{h}(p),\qquad
    K^{h'}(gq)=\rho_h(g)K^{h}(q).
\end{equation}
Their dot product is therefore unchanged:
\begin{align}
    \left\langle Q^{h'}(gp),K^{h'}(gq)\right\rangle
    & =Q^{h}(p)^{\top}\rho_h(g)^{\top}\rho_h(g)K^{h}(q)
    \\
    & =\left\langle Q^{h}(p),K^{h}(q)\right\rangle.
    \label{eq:invariant_dot_product}
\end{align}
The attention coefficients are consequently permuted consistently with the transformed spatial positions. Since the values transform through the same representation, the output satisfies
\begin{equation}
    Y^{h'}(gp)=\rho_h(g)Y^{h}(p).
\end{equation}
Thus,
\begin{equation}
    \operatorname{wG-CSA}(T_gx)=T_g\operatorname{wG-CSA}(x),
    \label{eq:wgcsa_equivariance}
\end{equation}
provided that the transformation maps complete windows to complete windows:
\begin{equation}
    W(gp)=gW(p).
    \label{eq:window_condition}
\end{equation}
This condition is satisfied by square windows under $90^{\circ}$ rotations and axis-aligned reflections when the square feature-map dimensions are divisible by the window size. 

\subsection{Quantitative Analysis}
We also evaluated the architectural equivariance of  REViT-v2s on the ImageNet validation set through equivariance error analysis.

\paragraph{Definition:}
For a feature map $f(\cdot)$, the equivariance error is defined as
\begin{equation}
\label{eq:app_eqerr}
\mathcal{E}_{\mathrm{eq}} =
\left\lVert
f(g \cdot x) - g \cdot f(x)
\right\rVert_{1},
\end{equation}
where $\lVert \cdot \rVert_{1}$ denotes the mean absolute difference over all spatial locations, channels, and batch elements. 

This metric is evaluated at two stages: 1) after the lifting layer (\emph{lifting equivariance error}) and 2) at the final feature representation after all the G-CSA transformer blocks and immediately preceding the classification head (\emph{pre-class equivariance error}).  In this setting, no trained checkpoint is loaded; the model is randomly initialized, so the experiment isolates equivariance induced by the network design rather than invariance learned through optimization. 1024 images were taken from ImageNet validation set and processed with standard ImageNet evaluation transforms: resize to 256, center cropping to $224 x 224$, and normalization with ImageNet mean and standard deviation.

For each sampled image and each non-identity element of the $p4$, $p4m$, and $p8$ groups, we compared the network response to the transformed input against the transformed response to the original input. Equivariance was measured at the stem output and immediately before the classification head (after all G-CSA blocks). Table \ref{tab:eq_summary_imagenet} summarizes the results of these equivariance tests.
For the $p8$ group, errors accumulate starting at the lifting layer, which we attribute to interpolation artifacts introduced by rotations at angles such as $45^\circ$ and $135^\circ$, where pixel locations fall between the original grid. This results in approximation error at the input level.

\begin{table}[tb]
\centering
\scriptsize
\caption{Equivariance error analysis}
\label{tab:eq_summary_imagenet}
\begin{tabular}{|c|c|c|}
\hline
 \textbf{Group} &  \textbf{Lifting  Stem Equivariance Err.$\downarrow$} & \textbf{Pre-Class Equivariance Err.$\downarrow$}  \\
\hline
$p4$   & 0.000142 $\pm$ 0.000017 & 0.00218 $\pm$ 0.000471  \\
$p4m$   & 0.001316 $\pm$ 0.000614 & 0.00008 $\pm$ 0.000034  \\
$p8$  & 0.001474$ \pm$ 0.000366 & 0.000102 $\pm$ 0.000013  \\
\hline
\end{tabular}
\end{table}

\section{Training Setup}
\label{app:revit_in}
REViT-v2 was trained on ImageNet using distributed data parallelism on four Nvidia GeForce RTX 4090 GPUs, with a per-GPU batch size of 128, giving an effective batch size of 512. We optimize the model for 300 epochs with AdamW, using an initial learning rate of 3e-4, weight decay of 0.05, and a learning-rate schedule consisting of 20 epochs of linear warmup followed by cosine decay. The model is instantiated with a window size of 7, and a $3x3$ equivariant kernel for the Q/K/V projections in the attention layers.

\end{document}